\documentclass[letterpaper]{article} 
\usepackage[draft]{aaai2026} 
\usepackage{times} 
\usepackage{helvet} 
\usepackage{courier} 
\usepackage[hyphens]{url} 
\usepackage{graphicx} 
\usepackage{natbib} 
\usepackage{caption} 
\usepackage{algorithm}
\usepackage{algorithmic}

\ifdefined
\fi

\title{Behavioral Controllability of Agentic Models for Information Extraction:\\From Fixed Workflows to Reflective Agents}

\author{Lujia Zhang, Xingzhou Chen, Hongwei Feng}
\affiliations{}

\begin{document}

\maketitle

\begin{abstract}
Large language model (LLM) agents are increasingly used for complex information-extraction tasks, yet it remains unclear whether agentic components such as reflection and memory lead to observable and controllable improvements over fixed LLM workflows. We study this question through conference-paper dataset extraction, where a system must identify datasets mentioned in scholarly PDFs and produce structured records. We compare a fixed workflow baseline with reflective agent variants and specify an optimized agent condition (S2) that extends the same task with richer PDF tools and dynamic tool selection. Our evaluation emphasizes process-level behavior---including tool execution, retries, reflection, memory use, runtime, and failure recovery---while treating extraction coverage and field completeness as secondary outcome measures. The paper characterizes when agentic mechanisms change system behavior, whether these changes improve task completion, and how the observed failure modes motivate an optimized agent design under the same evaluation harness.
\end{abstract}

\section{Introduction}
Large language models (LLMs) are increasingly embedded in systems that do more than produce a single response to a single prompt. Modern agentic systems connect LLMs to external tools, maintain intermediate state, decompose tasks into multiple actions, reflect on failed or incomplete outputs, and sometimes reuse prior experience through memory~\citep{yao2023react,schick2023toolformer,shinn2023reflexion,packer2023memgpt}. These capabilities are especially attractive for complex information-extraction tasks, where the required evidence may be distributed across long documents and where a single prompt is unlikely to capture all relevant context.

At the same time, it remains difficult to determine whether agentic components actually make such systems more controllable. Agent benchmarks such as AgentBench and WebArena show that tool-using agents can be evaluated in interactive environments, but they also reveal persistent gaps between autonomous execution and reliable task completion~\citep{liu2024agentbench,zhou2024webarena}. A fixed LLM workflow can already be a strong baseline: it can parse a document, apply carefully designed prompts, validate structured outputs, and save records in a reproducible order. By contrast, an agent may perform additional actions, retry uncertain steps, consult memory, or invoke reflection before continuing. These behaviors may improve coverage or recovery from errors, but they may also increase runtime, amplify mistakes, or create harder-to-debug execution traces. Thus, the key question is not simply whether an agent extracts more records, but whether its behavior changes in observable, configurable, and reproducible ways that are useful for the task.

We study this question in the setting of scholarly dataset extraction. Prior scientific text-mining resources and dataset-mention studies show that scholarly documents contain rich metadata, parsed full text, and ambiguous dataset mentions that require detection and linking~\citep{lo2020s2orc,pan2023dmdd,heddes2021dataset}. Given conference-paper PDFs and associated metadata, the system must identify datasets mentioned in each paper and produce structured dataset records containing fields such as dataset name, description, task or domain, paper reference, source link, and platform. This task is a realistic testbed for agentic information extraction. Dataset evidence can appear in abstracts, method sections, experiments, tables, captions, appendices, references, code links, or supplementary material. Mentions may be abbreviated, ambiguous, incomplete, or mixed with model names and benchmark suites. A workflow that follows a fixed sequence may miss evidence outside its initial extraction scope, while an agent has the opportunity to react to low-quality intermediate results by expanding the search, retrying with a different prompt, or using lessons from previous papers.

\begin{figure*}[t]
\centering
\includegraphics[width=\textwidth]{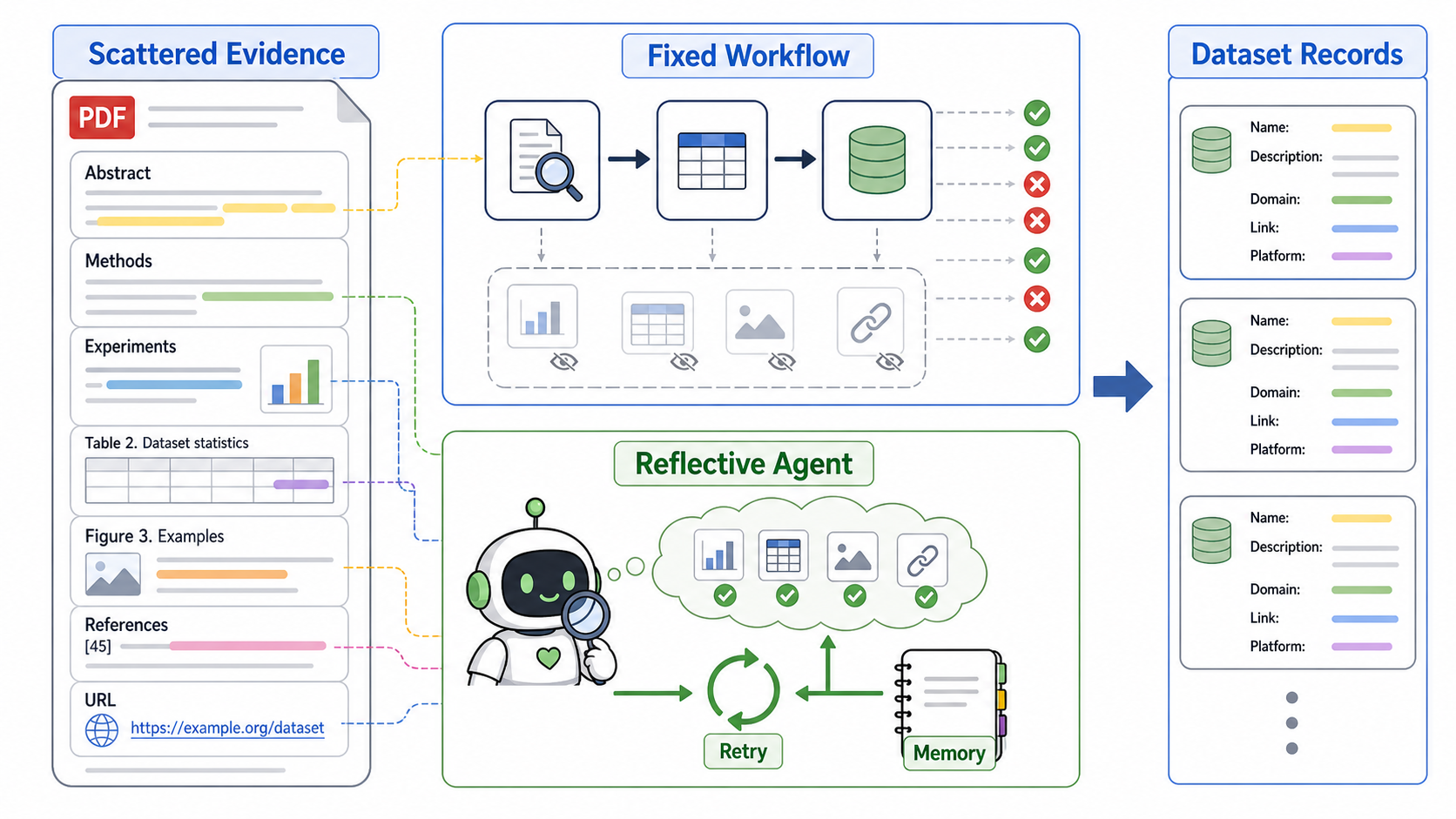}
\caption{Introductory motivation. Dataset evidence is scattered across paper sections, tables, figures, references, and URLs. A fixed workflow may miss evidence outside its initial extraction path, while a reflective agent can retry, consult memory, and make recovery behavior observable.}
\label{fig:intro-motivation}
\end{figure*}

We use this task to examine \emph{behavioral controllability}: the extent to which a system's decisions and adaptations are observable, configurable, reproducible, and comparable across experimental conditions. Observability requires traces that expose tool calls, intermediate outputs, reflection events, retries, memory accesses, errors, and stopping decisions. Configurability requires explicit parameters, such as retry limits, reflection settings, memory injection, and quality thresholds. Reproducibility requires recording prompts, model settings, tool behavior, and run metadata. Comparability requires that workflow and agentic systems operate on the same documents, output schema, and evaluation procedure.

Our empirical focus is therefore process-centric. We compare a fixed workflow baseline with agentic variants that implement ReAct-style tool use, reflection, retry policies, and memory, and we define an optimized S2 agent with twelve atomic tools and dynamic planning over the same corpus and schema. We measure final outputs, such as the number of dataset records and field completeness, but we treat these as only part of the evidence. The primary evidence comes from execution traces: action sequences, tool usage, reflection frequency, retry behavior, memory injection, failure recovery, runtime, and reproducibility artifacts.

This framing leads to three research questions:
\begin{itemize}
    \item \textbf{RQ1:} How does the decision process of an agent differ from that of a fixed workflow on the same extraction task?
    \item \textbf{RQ2:} Do reflection and memory produce measurable changes in tool use, retries, error recovery, and extraction outcomes?
    \item \textbf{RQ3:} What system-level optimizations are suggested by the observed behavior of reflective extraction agents?
\end{itemize}

The paper makes the following contributions:
\begin{itemize}
    \item We present a controlled comparison of fixed-workflow and agentic dataset-extraction systems under a shared corpus, output schema, and execution harness.
    \item We propose a process-centric evaluation framework that measures observable agent behavior in addition to final extraction outcomes.
    \item We analyze how reflection, memory, quality-aware retries, and tool policies affect extraction traces, recovery behavior, runtime, and result sets.
    \item We specify an S2 optimized agent with twelve atomic tools and dynamic planning, and define its evaluation protocol alongside empirical results for S0--S1b under the same corpus, schema, and logging harness.
    \item We derive practical design lessons for building scholarly extraction agents whose behavior is not only capable, but also inspectable, tunable, and reproducible.
\end{itemize}

\section{Related Work}
\subsection{LLM Agents and Tool Use}
A growing line of work treats LLMs as controllers that interleave language-based reasoning with external actions. ReAct~\citep{yao2023react} generates reasoning traces and tool calls in an interleaved loop, improving interpretability and reducing error propagation compared with reasoning-only or action-only baselines. Toolformer~\citep{schick2023toolformer} shows that models can learn when and how to invoke APIs through self-supervised filtering of candidate tool calls, without task-specific fine-tuning. These systems differ from fixed extraction pipelines in that action selection can depend on intermediate observations rather than a predetermined stage order. Our S1 agent adopts the ReAct-style pattern---observe, think, act, and reflect---while S0 remains a fixed workflow over the same tools and schema. Benchmarks such as AgentBench~\citep{liu2024agentbench} and WebArena~\citep{zhou2024webarena} evaluate autonomous agents in multi-environment and realistic web settings, but primarily report end-task success rather than the internal decision traces that motivate our controllability analysis.

\subsection{Reflection and Self-Correction}
Reflection and iterative refinement aim to improve outputs after an initial generation. Reflexion~\citep{shinn2023reflexion} stores verbal self-critiques in episodic memory and reuses them across trials without weight updates, yielding large gains on coding and decision-making tasks. Self-Refine~\citep{madaan2023selfrefine} alternates feedback and refinement steps with the same LLM across diverse generation tasks. Our S1a/S1b conditions instantiate a related pattern through rule-based and LLM-based reflection over tool outputs, coupled with quality-aware retries. However, \citet{huang2024selfcorrect} argue that intrinsic self-correction---correcting responses using only the model's own judgment and without reliable external feedback---often fails on reasoning tasks and can even degrade performance; reported gains frequently depend on oracle labels or stronger prompts in the feedback step. This caution is consistent with our empirical finding that LLM reflection changes process signals substantially while producing only modest output gains under a fixed retry budget.

\subsection{Memory and Experience Reuse}
Long-horizon agents require mechanisms to retain and retrieve past context. Generative agents~\citep{park2023generative} maintain a natural-language memory stream, periodically synthesize higher-level reflections, and retrieve memories to plan social behavior in an interactive sandbox. MemGPT~\citep{packer2023memgpt} applies an operating-system-inspired virtual context manager that pages information between working context and external storage, enabling document analysis and multi-session chat beyond a model's native context window. Our agent stores short- and long-term extraction experiences and optionally injects retrieved memory into prompts, which is closer to Reflexion-style episodic reuse than to MemGPT's paging architecture. Both prior lines of work highlight risks from stale, irrelevant, or incorrect memories; our S2 design therefore includes memory filtering and compression as explicit targets.

\subsection{Scholarly Information Extraction}
Scientific text mining has long relied on structured corpora and specialized mention detectors. S2ORC~\citep{lo2020s2orc} provides large-scale scholarly metadata, parsed full text, and linked inline citations, figures, and tables---infrastructure that motivates richer PDF tools than whole-document summarization alone. For dataset-centric extraction, DMDD~\citep{pan2023dmdd} offers the largest public corpus for dataset mention detection and linking, with weakly supervised in-text spans and a manually annotated evaluation set; it highlights ambiguity between dataset, method, and task names and the need for entity linking. Earlier work by \citet{heddes2021dataset} frames dataset-name recognition as NER over conference-paper sentences and releases a gold-standard annotation set with long, enumeration-heavy tag sequences. These studies focus on mention detection accuracy and corpus construction, whereas our task additionally requires structured record generation, link grounding, and agentic recovery when initial extraction is incomplete. PDF layout, tables, captions, references, and external URLs remain practical bottlenecks aligned with the evidence-grounding gaps observed in our case studies.

\subsection{Agent Evaluation and Controllability}
Agent benchmarks increasingly stress long-horizon, realistic tasks. AgentBench~\citep{liu2024agentbench} evaluates LLM agents across eight interactive environments with comparable success metrics, revealing large capability gaps across models and settings. WebArena~\citep{zhou2024webarena} provides reproducible, fully functional websites and programmatic validators for functional task correctness, showing that even strong LLM agents remain far below human performance on everyday web tasks. These benchmarks are essential for measuring what an agent accomplishes, but they offer limited visibility into retries, tool-selection policies, memory injections, and failure recovery. Our work complements outcome-oriented evaluation with a process-centric harness that logs action traces, reflection events, retry decisions, and run manifests under controlled interventions (S0 versus S1a/S1b versus S2). We argue that behavioral controllability---observability, configurability, reproducibility, and comparability---is itself an evaluation target for scholarly extraction agents, not only final record counts.

\section{Task and Experimental Framework}
\subsection{Task Definition}
The task takes as input a set of conference-paper PDFs together with available paper metadata, such as title, venue, year, authors, and source URL. The output is a JSONL file containing zero or more dataset records for each processed paper. Each record includes a dataset identifier, dataset name, a free-text dataset description, structured type/domain/field labels, paper-reference metadata, a dataset link when discovered, and a platform label such as GitHub, HuggingFace, OpenML, or unspecified. The same output schema is used for both the workflow baseline and the agentic systems so that differences can be attributed to system behavior rather than to downstream formatting.

The unit of analysis is multi-level. At the outcome level, we count final dataset records, unique papers covered, field completeness, link availability, and result-set overlap. At the process level, we analyze PDF parsing, LLM extraction calls, reflection events, retries, memory injection, tool execution statistics, and run metadata. This distinction is important because the project is not designed only as a dataset-construction pipeline; it is also an evaluation harness for observing how agentic mechanisms alter the behavior of an information-extraction system.

\subsection{Experimental Conditions}
We compare four conditions over the same NeurIPS 2024 extraction setting:
\begin{itemize}
    \item \textbf{S0 Workflow:} a fixed extraction workflow implemented in \texttt{main\_neurips.py}. It follows a predetermined sequence: download or load papers, parse PDFs, extract paper metadata, extract dataset names, extract dataset details, and append records to the output JSONL file. It has no reflection, no memory, and no agent-level retry policy.
    \item \textbf{S1a Agent with Rule Reflection:} an agentic system implemented in \texttt{main\_agent.py}. It uses a ReAct-style controller, tool manager, rule-based reflection, quality-aware retry, and memory injection, with LLM-based reflection disabled.
    \item \textbf{S1b Agent with LLM Reflection:} the same agentic harness as S1a, but with LLM-based reflection enabled in addition to rule reflection.
    \item \textbf{S2 Optimized Agent:} an extended agent implemented in \texttt{main\_agent\_s2.py}. It retains the S1 reflection, memory, and retry framework but registers twelve atomic tools (PDF retrieval, table and reference search, LLM extraction, link search, evidence verification, and record normalization) and selects the next tool dynamically from document state rather than following a fixed high-level plan.
\end{itemize}
Across conditions, we hold fixed the source corpus, target schema, underlying model family, maximum number of papers, and maximum number of datasets recorded per paper. The principal controlled interventions are therefore the extraction harness (fixed workflow versus reflective agent versus optimized agent) and, within the agent settings, the reflection and planning policies.

\subsection{Behavioral Controllability}
We operationalize behavioral controllability through four properties. First, \textbf{observability}: the agent logs observations, selected actions, tool results, reflection scores, retry decisions, and memory usage. Second, \textbf{configurability}: behavior is governed by explicit run controls for corpus size, per-paper output caps, retry budget, quality threshold, memory injection, and reflection mode. Third, \textbf{reproducibility}: each completed run records its condition, anonymized environment snapshot, artifact identifiers, and metrics file. Fourth, \textbf{comparability}: workflow and agent conditions process the same task and write structurally comparable JSONL outputs.

\section{System Design}
\subsection{Workflow Baseline}
The workflow baseline is a practical non-agentic extraction system. For each paper, it parses the PDF with a fixed parser, extracts a bounded text summary and URLs, prompts the LLM for paper metadata, prompts again for dataset names, and then prompts once per dataset name for structured details. The result is appended immediately to a JSONL output file, which supports long-running batch processing and partial recovery if a run is interrupted.

This baseline is strong enough to be meaningful because it uses the same PDF parser, prompt templates, LLM backend, and output schema as the agent. Its limitation is not that it lacks LLM capability, but that its action order is predetermined. If the extracted dataset list is incomplete, if a paper uses unusual dataset terminology, or if relevant evidence appears outside the initially summarized text, the workflow does not have an internal mechanism for reflecting on the deficiency, expanding the search, or learning from previous papers.

\subsection{Base Agent Architecture}
The agent system is organized around an \texttt{AgentController}. The controller maintains the current goal, plan, retry counts, and shared context for a paper. Its main loop follows the pattern Observe, Think, Act, Reflect, Learn, and Adjust. The current implementation uses a fixed high-level plan---parse the PDF, extract paper metadata, and extract dataset names---followed by a per-dataset detail extraction loop. Although the plan is fixed, each step is represented as an explicit action with parameters, result metadata, reflection, and memory update, making the behavior traceable and configurable.

Tools are managed by \texttt{ToolManager}. The registered tools include PDF parsing, paper-metadata extraction, dataset-name extraction, and dataset-detail extraction. LLM calls are routed through a shared client that can use an OpenAI-compatible API endpoint; the reported experiments use a vLLM deployment of \texttt{openPangu-Embedded-7B}, a 7B-parameter Pangu Embedded reasoner~\citep{chen2025pangu}. This design keeps the task interface stable while allowing the model backend and control policy to vary.

The reflection module evaluates each action result. Rule reflection assigns a quality score and identifies issues such as empty outputs, low-quality dataset extraction, or missing expected fields. If the score falls below the configured quality threshold and a diagnosable issue is found, the agent can retry the current step within the fixed retry budget. In S1b, LLM reflection is additionally enabled to provide a deeper diagnosis, although this increases runtime and produces more reflection events.

The memory module stores short-term session experiences and significant long-term experiences. When enabled, retrieved memory is injected into extraction prompts through a memory block. The design goal is to make previous successes and failures available to later papers without changing the output schema. This also creates a process-level signal: memory accesses and injections can be counted and inspected as part of controllability analysis.

\subsection{S2 Optimized Agent}
The S2 condition extends the S1 agent into an optimized extraction system implemented in \texttt{main\_agent\_s2.py}. It targets the same JSONL schema and logging harness, but replaces the fixed high-level plan with a controllable Observe--Plan--Act--Reflect--Learn--Adjust loop. Twelve atomic tools are registered in three groups: six PDF tools (parse, page extraction, section retrieval, table extraction, keyword search, and reference extraction), three LLM tools (metadata, dataset-name, and per-dataset detail extraction), and three grounding tools (link search, evidence verification, and record normalization). A planner selects the next tool from document state, pending datasets, planner milestones, and prior failures, combining hard constraints with optional LLM-based planning; when planning is unavailable or invalid, a deterministic fallback order applies. The controller tracks evidence use and stuck datasets, injects link hints into detail prompts, and applies S1-style reflection to tool outputs. Verification and normalization run only after per-dataset detail extraction. S2 is designed to test whether richer PDF access and state-dependent tool selection improve retry productivity, link grounding, and trace-level controllability under the same corpus and output budget as S0--S1b.

\begin{figure*}[t]
\centering
\includegraphics[width=\textwidth]{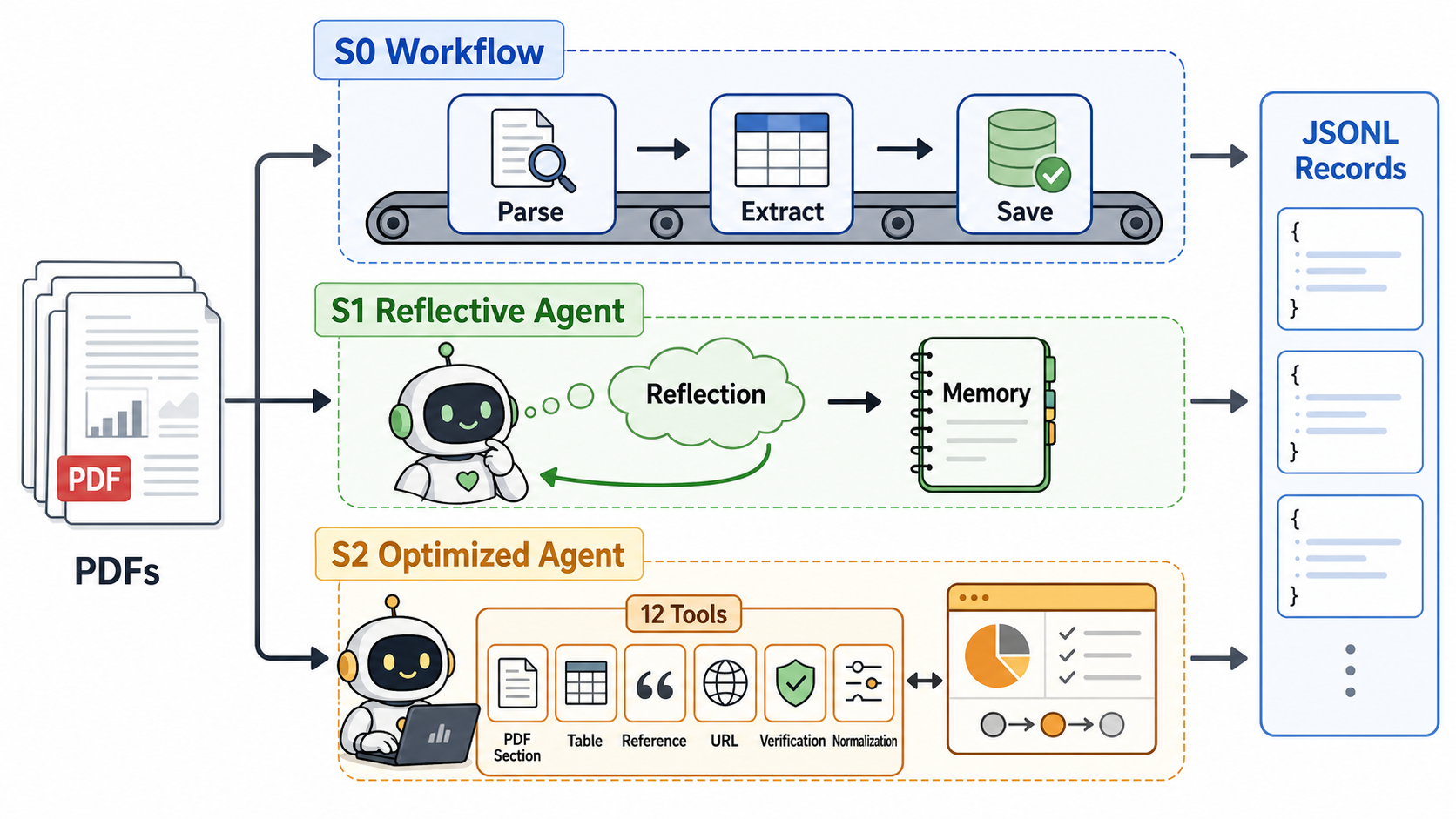}
\caption{System overview. S0 uses a fixed extraction sequence, S1 wraps the same task with observable agent actions, reflection, retries, and memory, and S2 specifies an optimized state-aware agent with expanded PDF and grounding tools.}
\label{fig:method-overview}
\end{figure*}

Figure~\ref{fig:method-overview} summarizes the architectural progression from the fixed workflow to the reflective agent and the optimized S2 design.

\subsection{Execution Trace and Logging}
The project records both outcome artifacts and process artifacts. Outcome artifacts include final JSONL files and per-run result copies. Process artifacts include logs, anonymized environment snapshots, manifests, PID files, and metric summaries. Each run manifest records the condition, system identifier, start time, model backend, model name, paper limit, dataset-per-paper limit, retry threshold, memory setting, reflection setting, and opaque artifact identifiers for outputs and logs. The summarization script converts these artifacts into metrics such as JSONL line count, unique-paper count, datasets per paper, field non-empty rates, log length, reflection mentions, observation mentions, and retry hints.

\section{Experimental Setup}
\subsection{Corpus}
The experiments use NeurIPS 2024 papers as the extraction corpus. Each condition processes at most 50 papers. The downloader streams papers and associated metadata, while the parser extracts text and URLs from the PDFs. The latest result package contains three June 6 runs: S0 Workflow, S1a Agent with rule reflection, and S1b Agent with LLM reflection. Each paper can contribute up to five dataset records, a practical cap that keeps systems comparable under a fixed output budget.

We do not assume a complete human-annotated ground truth for all dataset mentions. Consequently, the current evaluation treats record count, field completeness, and pairwise differences as task-completion indicators rather than definitive precision or recall. The main target of analysis is the behavioral trace produced by each system.

\subsection{Implementation Details}
All June 6 runs use the same vLLM backend serving \texttt{openPangu-Embedded-7B}~\citep{chen2025pangu} through an OpenAI-compatible API, with a 900-second request timeout. The paper corpus is stored in an anonymized cache, and no machine-specific storage location is required for reproducing the analysis. Shared controls fix the year, paper limit, per-paper record cap, retry budget, and quality threshold; the concrete values are listed in Appendix Table~\ref{tab:repro-env}. Memory injection is enabled for both agent runs. S1a disables LLM reflection, while S1b enables it.

All conditions are launched through the same experiment runner, which creates a run directory and records manifests, outputs, and metrics for each condition. Because the system relies on long LLM calls and PDF-processing side effects, the current evaluation reports single-run descriptive results rather than repeated-run statistical estimates.

\subsection{S2 Experimental Design}
S2 is configured to use the same NeurIPS 2024 corpus, paper limit, per-paper cap, retry budget, and vLLM-served \texttt{openPangu-Embedded-7B} backend~\citep{chen2025pangu} as S0--S1b. The planned comparison contrasts S2 against S1a and S1b on outcome metrics (record count, link rate, field completeness), process metrics (tool diversity, section/table/reference tool usage, reflection and retry counts), and controllability metrics (logged planner decisions, stuck-dataset recovery, and run manifests). Since the available archived result package contains completed S0--S1b runs but not a completed S2 run, we report S2 as a specified optimized-agent design and reserve its numerical evaluation for follow-up experiments.

\subsection{Evaluation Metrics}
\paragraph{Process Metrics.}
Process metrics describe what the system did before producing final records. For the workflow, the available process signals are primarily paper-processing logs and output counts. For the agent, additional signals include observation events, reflection mentions, retry hints, memory injection, and tool-level execution statistics. These metrics are used to determine whether agentic mechanisms produce observable behavioral changes, not merely whether the final JSONL file is longer.

\paragraph{Outcome Metrics.}
Outcome metrics include total dataset-record count, unique papers represented in the output, datasets per paper, and non-empty rates for selected fields. The current summaries report link rate, platform rate, and content rate. These values are interpreted as auxiliary task-completion metrics. Since the project does not yet include a complete manually verified gold standard, we avoid claiming absolute extraction accuracy from these metrics alone.

\paragraph{Controllability Metrics.}
Controllability metrics are derived from whether decisions are logged, whether behavior is governed by explicit parameters, and whether runs can be reconstructed from manifests and metrics. For example, the difference between S1a and S1b is controlled by a single reflection flag while the task, model, paper limit, retry budget, memory setting, and output schema remain fixed.

\paragraph{Manual Validation Protocol.}
Because the current package does not include a complete human-labeled gold standard, we define a manual audit protocol for the follow-up submission version. For each condition, annotators should sample matched and condition-specific \((\mathrm{paper}, \mathrm{dataset})\) pairs, verify whether the dataset is genuinely mentioned in the paper, check whether the record describes a dataset rather than a method or task, and mark whether the link is supported by text, references, tables, captions, or external resource pages. The audit should report precision for sampled records, link-grounding precision, missing-field rates, and disagreement cases. This protocol separates the present process-controllability claims from future accuracy claims.

\subsection{Ablations and Statistical Analysis}
The current runs implement an initial ablation over reflection: S1a disables LLM reflection while retaining rule reflection, memory injection, and retry; S1b enables LLM reflection. The workflow baseline removes the agent harness entirely. Additional ablations---such as disabling memory injection, disabling quality-aware retries, or introducing stronger memory filtering---are left for follow-up runs. Because the present result package contains one run per condition, the analysis is descriptive and paired by corpus rather than inferential.

S2 ablations should isolate dynamic planning, expanded PDF tools, and memory-in-planner settings while holding the corpus, schema, and output cap fixed. These ablations are treated as future validation rather than as part of the completed June 6 result package.

\section{Results}
\subsection{Overall System Comparison}
Table~\ref{tab:main-results} summarizes the latest June 6 runs using the same statistics computed directly from each archived \texttt{results.jsonl}. The workflow baseline produces 158 records from 42 unique papers, with 3.76 datasets per paper. The agent outputs increase the number of records: S1a produces 165 records from 43 papers, and S1b produces 168 records from 43 papers. Content fields are complete in all reported summaries. Link discovery remains relatively low across systems, indicating that source-link extraction is a persistent bottleneck independent of the agent harness.

\begin{table}[t]
\centering
\caption{Latest NeurIPS 2024 extraction results. D/Paper denotes datasets per covered paper.}
\label{tab:main-results}
\resizebox{\columnwidth}{!}{%
\begin{tabular}{lrrrr}
\hline
Condition & Rec. & Papers & D/Paper & Link \\
\hline
S0 Workflow & 158 & 42 & 3.76 & 0.190 \\
S1a Rule Reflection & 165 & 43 & 3.84 & 0.164 \\
S1b LLM Reflection & 168 & 43 & 3.91 & 0.185 \\
\hline
\end{tabular}
}
\end{table}

\begin{figure*}[t]
\centering
\setlength{\unitlength}{0.86mm}
\begin{picture}(200,70)
\put(7,60){\small (a) Extraction coverage}
\put(18,12){\vector(0,1){42}}
\put(18,12){\vector(1,0){70}}
\put(13,10){\scriptsize 40}
\put(10,30){\scriptsize 100}
\put(10,49){\scriptsize 160}
\put(18,30){\line(1,0){66}}
\put(18,49){\line(1,0){66}}
\put(24,4){\scriptsize S0}
\put(48,4){\scriptsize S1a}
\put(72,4){\scriptsize S1b}
\put(28,49){\circle*{2}}
\put(52,51){\circle*{2}}
\put(76,52){\circle*{2}}
\put(28,49){\line(6,1){24}}
\put(52,51){\line(6,1){24}}
\put(25,53){\scriptsize 158}
\put(49,55){\scriptsize 165}
\put(73,56){\scriptsize 168}
\put(27,12){\framebox(2,2){}}
\put(51,12){\framebox(2,2){}}
\put(75,12){\framebox(2,2){}}
\put(28,13){\line(1,0){24}}
\put(52,13){\line(1,0){24}}
\put(24,17){\scriptsize 42}
\put(49,17){\scriptsize 43}
\put(73,17){\scriptsize 43}
\put(24,63){\scriptsize Filled dots: records \quad Squares: papers}

\put(111,60){\small (b) Density and grounding}
\put(122,12){\vector(0,1){42}}
\put(122,12){\vector(1,0){70}}
\put(110,12){\scriptsize 3.70}
\put(110,31){\scriptsize 3.82}
\put(110,50){\scriptsize 3.94}
\put(122,31){\line(1,0){66}}
\put(122,50){\line(1,0){66}}
\put(128,4){\scriptsize S0}
\put(151,4){\scriptsize S1a}
\put(176,4){\scriptsize S1b}
\put(132,21){\circle*{2}}
\put(156,34){\circle*{2}}
\put(180,45){\circle*{2}}
\put(132,21){\line(2,1){24}}
\put(156,34){\line(2,1){24}}
\put(128,25){\scriptsize 3.76}
\put(151,38){\scriptsize 3.84}
\put(175,49){\scriptsize 3.91}
\put(132,47){\makebox(0,0){\scriptsize $\triangle$}}
\put(156,19){\makebox(0,0){\scriptsize $\triangle$}}
\put(180,43){\makebox(0,0){\scriptsize $\triangle$}}
\put(132,47){\line(1,-1){24}}
\put(156,19){\line(1,1){24}}
\put(127,51){\scriptsize 19.0\%}
\put(150,13){\scriptsize 16.4\%}
\put(174,37){\scriptsize 18.5\%}
\put(128,63){\scriptsize Filled dots: datasets/paper \quad Triangles: link rate}
\end{picture}
\caption{Outcome trends across the completed June 6 conditions. Agentic variants increase record count and datasets per paper relative to the fixed workflow, while link grounding remains low and non-monotonic.}
\label{fig:result-trends}
\end{figure*}
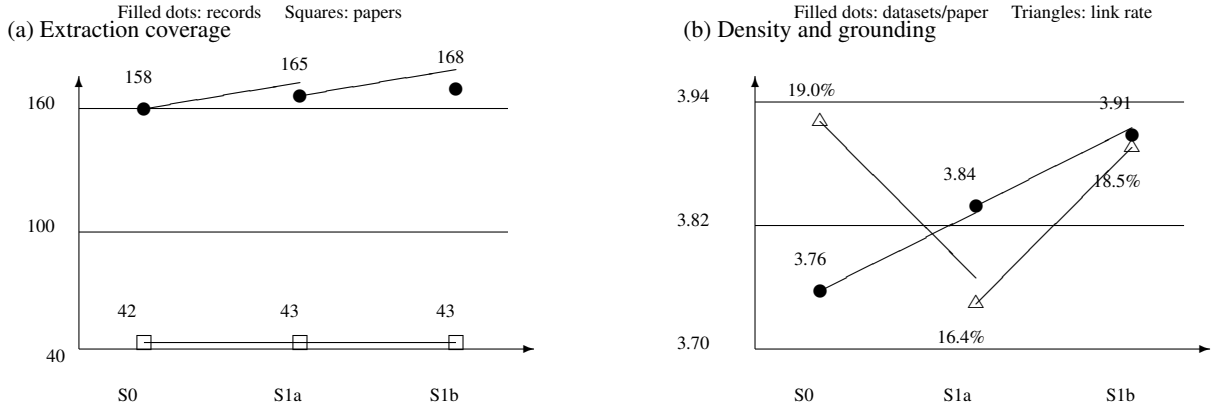

These results suggest that the agent harness increases extraction coverage under the same paper and per-paper output budget. Figure~\ref{fig:result-trends} visualizes the same pattern: record count and datasets per paper rise from S0 to S1b, whereas link rate remains low and non-monotonic. However, the magnitude is modest: S1b produces ten more records than S0. The stronger difference is behavioral. The agent logs contain observation, reflection, and retry signals that are absent from the workflow baseline. Thus, the main empirical finding is not only that the agent extracts more records, but that its behavior is more inspectable and parameterized.

\subsection{Workflow Versus Base Agent}
Compared with S0, the agent conditions introduce explicit intermediate states. Each paper is processed through actions whose results are reflected on and stored as experiences. If dataset extraction appears low quality, the agent can retry with expanded context and merge newly found dataset names. This changes the extraction process from a one-pass sequence into a quality-aware loop.

The outcome difference is visible in the June 6 outputs: S0 records 158 datasets, whereas the agent outputs record 165--168 datasets. Several high-density papers reach the per-paper cap of five records, including papers on surgical video-language pretraining, anomaly detection, off-policy evaluation, multimodal image fusion, and anchoring for vision models. These cases illustrate why a fixed cap is useful: without it, a small number of papers with many benchmark datasets could dominate aggregate comparisons.

The agent's additional behavior also carries cost. The S1 metrics report a much larger log trace than S0 and thousands of reflection- or retry-related mentions. This indicates substantially more internal activity. Without a complete gold standard, we cannot claim that every additional record is correct or useful. The result should be interpreted as increased coverage and increased observability, accompanied by increased execution complexity.

\subsection{Rule Reflection Versus LLM Reflection}
S1a and S1b isolate the effect of enabling LLM reflection on top of the same agent harness. In the latest archived outputs, S1a produces 165 records and S1b produces 168 records. The small record-count difference suggests that LLM reflection does not dramatically change aggregate extraction coverage under the present configuration. Its main expected effect is instead on the process trace: deeper diagnostic reflection, more opportunities to justify retries, and potentially better explanations of failure modes.

The archived logs confirm that the agent harness produces the intended process signals. The June 6 S1 logs contain observation markers, reflection markers, retry decisions, memory retrieval, memory injection, tool-call timing, per-paper session summaries, and final tool statistics. Table~\ref{tab:s1-process} summarizes log-derived process signals from the same June 6 run package as Table~\ref{tab:main-results}. Because S1a and S1b append to a shared agent log, mention counts reflect cumulative agent activity through the June 6 S1b run completion; run-isolated outcome differences remain visible in Table~\ref{tab:main-results} (165 versus 168 records). Relative to S0 (7,084 log lines), the agent log is much larger (50,189 lines), indicating substantially more internal activity even when additional records are modest.

\begin{table}[t]
\centering
\caption{Log-derived process signals for the June 6 runs. S1 mention counts come from the shared agent log through S1b completion.}
\label{tab:s1-process}
\resizebox{\columnwidth}{!}{%
\begin{tabular}{lrrr}
\hline
Metric & S0 & S1a & S1b \\
\hline
Log lines & 7,084 & 50,189 & 50,189 \\
Reflection mentions & --- & 1,967 & 1,967 \\
Observation mentions & --- & 1,388 & 1,388 \\
Retry mentions & --- & 2,222 & 2,222 \\
\hline
\end{tabular}
}
\end{table}

The output-level comparison shows that LLM reflection adds only three additional records over rule reflection in the June 6 run package. The process-level comparison shows much larger agent logs than S0 and thousands of reflection- and retry-related mentions. This motivates a process-centric reading: rule reflection already changes the extraction harness relative to S0, and LLM reflection should be evaluated by whether it produces more useful diagnoses and more productive retries---as illustrated in the case studies---not only by whether it increases the number of extracted records.

\subsection{Ablation Results}
The implemented ablation separates three levels of control: no agent harness (S0), agent with rule reflection (S1a), and agent with rule plus LLM reflection (S1b). The largest structural difference is between S0 and the two agent conditions, because only the agent records reflection, retry, and memory behavior. The smaller difference between S1a and S1b suggests that rule-based reflection and retry already account for much of the measurable output change in this configuration. LLM reflection may still be valuable for qualitative diagnosis, but the present aggregate results do not show a large record-count gain.

\subsection{Case Studies}
We identify three trace-backed case studies from the archived agent logs and result sets.

\paragraph{Case 1: Normal ReAct execution with memory creation.}
For ``Procedure-Aware Surgical Video-language Pretraining with Hierarchical Knowledge Augmentation,'' the S1a log records the full ReAct sequence. The agent parses a 32-page PDF, extracts 99,901 characters, identifies 80 dataset-related sentences and one URL, and then executes metadata extraction and dataset-name extraction. The dataset-name step receives a quality score of 1.00, stores the experience in long-term memory under the \texttt{extract\_datasets} pattern, and proceeds to per-dataset detail extraction. The initial extraction finds seven dataset names, after which the run-level cap limits the saved records to three: SVL, Cholec80, and AutoLaparo. This case illustrates the basic behavior of the harness: every stage is logged as an action, scored by reflection, and converted into memory.

\paragraph{Case 2: Rule reflection triggers bounded retry.}
For ``Span-Based Optimal Sample Complexity for Weakly Communicating and General Average Reward MDPs,'' the S1a log shows a low-quality dataset-extraction result. The agent assigns a quality score of 0.10, records the issue ``no datasets extracted,'' suggests expanding the search scope or adjusting keywords, and retries the \texttt{extract\_datasets} action. The retry path re-parses the 50-page PDF, expands the search text, injects retrieved memory into the prompt, and repeats extraction. After the second retry again receives a quality score of 0.10, the agent reaches the retry limit and skips the step, saving zero datasets for that paper. This is a useful negative case: the agent does not silently fail, but it also does not recover. The trace exposes the failure cause and the stopping decision.

\paragraph{Case 3: LLM reflection provides diagnostic guidance.}
For ``Leveraging Catastrophic Forgetting to Develop Safe Diffusion Models against Malicious Finetuning,'' the S1b log shows LLM reflection changing the character of the trace. After repeated \texttt{extract\_datasets} attempts, the LLM-reflection step assigns a quality score of 0.78 and reports risks that are more specific than the rule-only messages: the extraction depends heavily on truncated abstract/introduction context, and the experimental setup section may contain missing datasets. It recommends prioritizing sections with keywords such as ``Experiments,'' ``Datasets,'' and ``Evaluation.'' The agent then proceeds to detail extraction for COCO-30K, I2P, and LAION-5B, injecting memory into each detail prompt. This case demonstrates the main value of LLM reflection in the current system: it produces actionable diagnosis and design evidence for S2, even when the current tool set cannot yet execute a true section-level retrieval policy.

Across these cases, link grounding remains the most visible weakness. Link rate stays below 20\% even though content descriptions are usually present. The traces suggest why: the current prompts can describe datasets from context, but URL grounding is treated as a field inside a broad extraction call rather than as a dedicated verification action. The S2 design addresses this gap by treating URL matching, citation search, and evidence verification as first-class tools; a follow-up run should test whether these tools change link rate and recovery behavior in practice.

\section{Analysis and Discussion}
\subsection{Answers to the Research Questions}
\textbf{RQ1.} The fixed workflow and the agent differ primarily in process structure. The workflow is a one-pass extraction pipeline with predictable stages and limited recovery. The agent exposes each step as an action with observation, reflection, learning, and adjustment. Even when the high-level plan remains fixed, the agent produces a richer behavioral trace and supports quality-triggered retries.

\textbf{RQ2.} Reflection and memory produce clear process-level changes: the agent logs reflection events, retry hints, and memory-related behavior that do not exist in S0. These changes correspond to a modest increase in output records in the latest runs. However, the available evidence is stronger for behavioral change than for absolute accuracy improvement, because no complete human-labeled gold standard is available.

\textbf{RQ3.} The S1 runs identify concrete optimization targets---stronger evidence grounding, cleaner source-link extraction, more selective reflection, richer PDF tools, and better memory filtering---that motivate the S2 optimized agent. The current evidence supports S2 as a design response to observed failure modes; its numerical evaluation remains future work.

\subsection{Cost--Benefit Trade-Offs}
The workflow remains attractive when the goal is cheap, predictable batch extraction and when a fixed prompt sequence is sufficient. It is simpler, easier to run, and easier to debug at the level of final outputs. The agent is more attractive when the corpus contains heterogeneous papers, many dataset mentions, ambiguous benchmark names, or failure modes that can be diagnosed from intermediate outputs.

The main trade-off is between behavioral richness and execution cost. Reflection and retry create more opportunities for recovery, but they also increase the number of model calls and the complexity of logs. LLM reflection in particular should be used selectively: the current aggregate output difference between S1a and S1b is small, so indiscriminate deep reflection may not be justified for every paper.

\subsection{Implications for Agent Harness Design}
The project suggests four design lessons. First, reflection should be tied to observable deficiencies, such as empty outputs, low dataset counts despite benchmark cues, or missing required fields. Second, memory should include provenance and quality control; otherwise, irrelevant prior experiences may become prompt noise. Third, better tools are useful only if the controller has policies for selecting them based on state. Fourth, controllability is a property of the whole harness: logging, configuration, run manifests, and output schemas matter as much as the underlying LLM.

\subsection{Limitations and Threats to Validity}
The study has several limitations. The corpus is limited to NeurIPS 2024 and a maximum of 50 processed papers per run. The reported experiments use a fixed vLLM deployment of \texttt{openPangu-Embedded-7B}~\citep{chen2025pangu}, so the findings may not transfer directly to other model families or serving stacks. The task lacks a complete manually verified gold standard, which limits claims about precision and recall. The results depend on PDF parsing quality and on whether dataset evidence appears in the extracted text window. The agent implementation also combines architectural changes with prompt and retry-policy changes, so some observed differences cannot be attributed to a single component. Finally, process metrics reveal what the harness logs, but they should not be interpreted as faithful access to the model's internal reasoning.

\section{Future Work}
\begin{itemize}
    \item \textbf{S2 Follow-Up Ablations:} After the main S2 run, evaluate planner-on versus planner-off settings, memory-in-planner variants, and selective LLM reflection to identify which optimized-agent components contribute most to link grounding and productive retries.
    \item \textbf{Toward Whole-Paper Understanding Agents:} Dataset extraction should eventually become one component of a broader scholarly-paper understanding agent. Such an agent would combine dataset extraction with method extraction, task and benchmark identification, code/resource discovery, contribution summarization, limitation analysis, and evidence-grounded paper comparison. In this direction, the present dataset miner serves as a controllable submodule: it provides structured tool traces, reflection signals, memory behavior, and reproducible outputs that can be integrated into a larger agent for understanding the full content of scientific papers.
\end{itemize}

\section{Conclusion}
We presented a controlled comparison between a fixed LLM workflow, reflective agentic systems, and an optimized S2 agent design for scholarly dataset extraction. On NeurIPS 2024 papers, the completed S1 agent conditions produce slightly more dataset records than the workflow baseline under the same paper limit and per-paper output cap. The S2 design translates observed S1 failure modes into concrete tool and planner requirements for future evaluation. More importantly, the agent conditions expose a richer decision process through observations, reflections, retries, memory injection, and run-level metadata.

The results support a process-centric view of agent evaluation. Agentic mechanisms should not be judged only by whether they increase final record counts. They should also be judged by whether their behavior is observable, configurable, reproducible, and meaningfully adaptive to task failures. In this project, reflection, memory, retry thresholds, and run manifests provide a practical foundation for studying behavioral controllability in LLM-based information extraction.

\appendix
\section{Reproducibility Details}
The experiments are launched through a common runner that dispatches the workflow, rule-reflection agent, LLM-reflection agent, or optimized-agent implementation according to the selected condition. Each completed run writes a manifest containing the anonymized run identifier, condition, system label, start time, environment snapshot, output artifact identifier, log artifact identifier, and metrics artifact identifier. The completed June 6 runs are referenced as \texttt{S0-June6}, \texttt{S1a-June6}, and \texttt{S1b-June6}; S2 uses the same manifest format for a follow-up optimized-agent run.

\begin{table}[!htbp]
\centering
\caption{Key environment settings for the June 6 runs.}
\label{tab:repro-env}
\resizebox{\columnwidth}{!}{%
\begin{tabular}{ll}
\hline
Variable & Value \\
\hline
\texttt{DATA\_CACHE} & anonymized corpus cache \\
\texttt{LLM\_PROVIDER} & \texttt{vllm} \\
\texttt{VLLM\_MODEL} & \texttt{openPangu-Embedded-7B} \\
\texttt{VLLM\_TIMEOUT} & \texttt{900} \\
\texttt{NEURIPS\_YEAR} & \texttt{2024} \\
\texttt{MAX\_PAPERS} & \texttt{50} \\
\texttt{MAX\_DATASETS\_PER\_PAPER} & \texttt{5} \\
\texttt{MAX\_RETRIES} & \texttt{2} \\
\texttt{QUALITY\_RETRY\_THRESHOLD} & \texttt{0.6} \\
\texttt{ENABLE\_MEMORY\_IN\_PROMPTS} & \texttt{true} \\
\hline
\end{tabular}
}
\end{table}

The output schema is a JSONL record with the following top-level fields: \texttt{dataset id}, \texttt{name}, \texttt{dataset describe}, \texttt{paper\_refs}, \texttt{dataset link}, and \texttt{platform}. The \texttt{dataset describe} object contains \texttt{content}, \texttt{type}, \texttt{domain}, and \texttt{fields}. The \texttt{paper\_refs} object contains paper title, authors, institutions, venue, year, URL, and \texttt{is\_fellow}. This schema is shared by S0, S1a, S1b, and S2.

The prompt file defines three main extraction prompts: paper metadata extraction, dataset-name extraction, and per-dataset detail extraction. The agent versions include a \texttt{\{memory\_block\}} insertion field, allowing retrieved experience to be injected into the metadata, dataset-name, and dataset-detail prompts when memory injection is enabled. The dataset-name prompt asks the model to return a JSON object of dataset names only, while the detail prompt asks for name, content, type, domain, fields, dataset link, and platform. JSON responses are parsed directly or extracted from fenced JSON blocks.

The default agent tool set contains four tools registered by \texttt{ToolManager}: PDF parsing, metadata extraction, dataset-name extraction, and dataset-detail extraction. The controller logs observation, reasoning, action, result, reflection, issue, suggestion, retry, and memory-related events. The summarization script \texttt{scripts/summarize\_run.py} computes line count, unique-paper count, datasets per paper, link rate, platform rate, content rate, top papers by record count, and log counters for reflection, observation, and retry mentions.

\section{Additional Results}
Pairwise result-set overlap is computed over \((\mathrm{paper\ title}, \mathrm{dataset\ name})\) pairs. S0 and S1a share 136 pairs; 22 appear only in S0 and 29 appear only in S1a. S0 and S1b share 138 pairs; 20 appear only in S0 and 30 appear only in S1b. S1a and S1b are much closer, sharing 161 pairs, with 4 only in S1a and 7 only in S1b.

\begin{table}[!htbp]
\centering
\caption{Pairwise result-set overlap over \((\mathrm{paper}, \mathrm{dataset})\) pairs.}
\label{tab:overlap}
\begin{tabular}{lrrr}
\hline
Comparison & Shared & Only A & Only B \\
\hline
S0 vs. S1a & 136 & 22 & 29 \\
S0 vs. S1b & 138 & 20 & 30 \\
S1a vs. S1b & 161 & 4 & 7 \\
\hline
\end{tabular}
\end{table}

Examples of records appearing in S1 but not S0 include KDDCup2k, MSNBC, NLTCS, Plants, Argoverse 2, KITTI, GBD13, and QM9. Examples appearing in S1b but not S1a include LassoBench, BAMBOO, Mixamo+, and CIMA. These examples should be treated as candidates for manual verification rather than as automatic proof of correctness.

\bibliography{references}

\end{document}